\pdfoutput=1

\documentclass[11pt]{article}

\usepackage[]{acl}

\usepackage{times}
\usepackage{latexsym}
\usepackage{bm}
\usepackage{hyperref}
\usepackage{color,colortbl}
\usepackage{framed}
\definecolor{shadecolor}{rgb}{0.92,0.92,0.92}
\usepackage{stfloats}
\usepackage{enumitem}
\usepackage{booktabs}
\usepackage{nccmath}
\usepackage{multirow}
\usepackage{subfigure}
\usepackage{latexsym}
\usepackage{todonotes}
\usepackage{makecell}
\usepackage{dashrule}
\usepackage{amssymb}
\usepackage{bbding}
\usepackage{arydshln}
\usepackage[T1]{fontenc}

\usepackage[utf8]{inputenc}

\usepackage{microtype}

\usepackage{inconsolata}
\usepackage{multirow}
\usepackage{subfigure}
\usepackage{makecell}

\title{Smooth Reward Tuning Improves Reasoning in Language Models}
\title{A Little Math: Smooth Reward Tuning Improves \\ Reasoning in Small Language Models}
\title{A Little Math: Reward Tuning Improves \\ Reasoning in Small Language Models}

\title{A Little Math around World: Multilingual Reasoning with Small Language Models}

\title{A Little Math: \\ Small Language Models are Great at  Multilingual Reasoning}

\title{A Little Math: \\Multilingual Instruction Tuning Improves Reasoning Consistency}

\title{\texttt{mCoT}: Multilingual Instruction Tuning for Reasoning Consistency in Language Models}

\author{Huiyuan Lai \and Malvina Nissim \\
    Center for Language and Cognition (CLCG) \\
    University of Groningen / The Netherlands\\
  \texttt{\{h.lai, m.nissim\}@rug.nl} \\
}

\begin{document}
\maketitle
\begin{abstract}
Large language models (LLMs) with Chain-of-thought (CoT) have recently emerged as a powerful technique for eliciting reasoning to improve various downstream tasks. As most research mainly focuses on English, with few explorations in a multilingual context, the question of how reliable this reasoning capability is in different languages is still open. To address it directly, we study multilingual reasoning consistency across multiple languages, using popular open-source LLMs. 
First, we compile the first large-scale multilingual math reasoning dataset, \texttt{mCoT-MATH}, covering eleven diverse languages. Then, we introduce multilingual CoT instruction tuning to boost reasoning capability across languages, thereby improving model consistency. While existing LLMs show substantial variation across the languages we consider, and especially low performance for lesser resourced languages, our 7B parameter model \texttt{mCoT} achieves impressive consistency across languages, and superior or comparable performance to close- and open-source models even of much larger sizes. 

\end{abstract}

\section{Introduction}


Recent progress on language models shows that they can achieve surprising performance on complex reasoning tasks in natural language processing (NLP), such as symbolic reasoning~\citep{wei2022chain, kojima2023large}, math word problem~\citep{cobbe-etal-2020-training, wei2022chain}, and commonsense reasoning~\citep{wei2022chain, kojima2023large}. Most of the research focuses on prompting large language models (LLMs), where the LLMs are conditioned on a few examples or instructions describing the target task~\citep{wei-2022-chain, fu-2023-complexity, zhou2023leasttomost, kojima2023large, zheng-2023-php}.

\begin{figure}[!t]
\centering
\includegraphics[scale=.55]{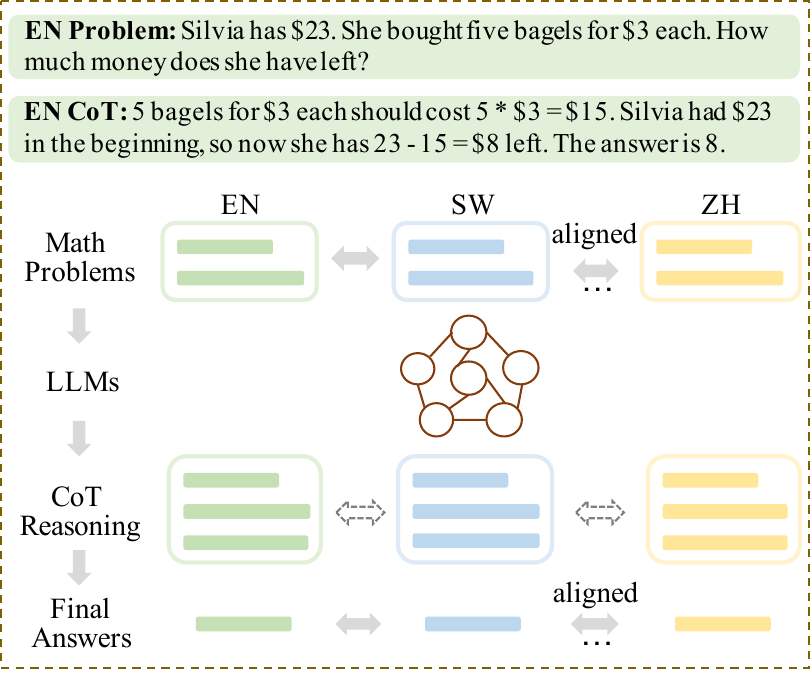}
\caption{Overview of multilingual reasoning; LLMs are expected to have consistent reasoning capabilities across different languages when given the same problem which has the same answer. Shown in picture are three example languages: English (EN), Swahili (SW), and Chinese (ZH). For EN, we show the problem formulation, and the Chain-of-Thought (CoT) reasoning.} 
\label{fig:mcot}
\end{figure}

While most previous works focus on reasoning with LLMs in English,~\citet{shi2023language} recently have extended it to a multilingual setting leveraging a few-shot prompting approach. However, performance for lesser resourced languages still lags behind, in a similar way as generalization of factual knowledge has been shown to vary widely across languages~\citep{fierro-sogaard-2022-factual, qi-etal-2023-cross}, mainly due to the fact that most languages are not well represented in LLMs. The work we present here has the twofold aim of (i) better understanding and evaluating the general reasoning capabilities of LLMs beyond just English, and (ii) providing lesser resourced languages with capable but manageable models which can be used for reasoning tasks.

To this end, we propose to measure reasoning consistency across multiple languages. As shown in Figure~\ref{fig:mcot}, LLMs are expected to produce logically similar reasoning solutions and consistent final results for inputs which are semantically equivalent but expressed in different languages. Based on our findings, we aim to enhance multilingual reasoning abilities through instruction tuning, yielding a model that can solve reasoning tasks in various languages, with similar reasoning capabilities across those languages.
Specifically, we focus on math word problems and empirically investigate the reasoning consistency of current open-source state-of-the-art LLMs across multiple languages. The model's reasoning consistency is evaluated toward the final answer (the final results should be the same across languages). On the basis of preliminary results showing substantial reasoning gaps between different languages, we propose a multilingual Chain-of-Thought reasoning (mCoT) framework using instruction tuning, which aims to boost reasoning capability across languages, thereby improving consistency. So, first we construct the multilingual instruction training dataset (mCoT-MATH) by automatically translating English data into multiple languages, and then we use it for LLM finetuning. 

In summary, our contributions are:\footnote{Data, code, and model are available at \url{https://github.com/laihuiyuan/mcot}.}

\begin{itemize}[itemsep=1.2pt, topsep=1.2pt, parsep=1.2pt]
\item We propose to study reasoning consistency of LLMs across different languages, providing insights into (the evaluation of) this ability of LLMs.

\item We compile and distribute mCoT-MATH, the first large-scale multilingual math CoT reasoning dataset containing around 6.3 million samples for 11 diverse languages.

\item Based on mCoT-MATH, we train and make available a 7B parameter model mCoT for multilingual math reasoning, which achieves impressive consistency across languages, and superior or comparable performance to close- and open-source models.

\end{itemize}

\section{Background}

\paragraph{Math Reasoning with LLMs}
In recent years, LLMs such as GPT-3~\citep{brown-etal-2020-language} have shown impressive performance in various NLP tasks. In particular, LLMs with CoT-based method exhibit the emergent ability to perform complex math reasoning tasks~\citep{wei-2022-chain, wang2023selfconsistency}. One popular line is prompt engineering, which aims to elicit reasoning capability of LLMs by exploring various prompts, such as basic CoT prompting~\citep{wei-2022-chain}, complex CoT~\citep{fu-2023-complexity}, auto-CoT~\citep{zhang2023automatic}, self-consistency CoT~\citep{wang2023selfconsistency}, multilingual CoT~\citep{shi2023language}, least-to-most prompting~\citep{zhou2023leasttomost}, progressive-hint prompting~\citep{zheng-2023-php}, residual connection prompting~\citep{jiang2023resprompt}, and using speciﬁc phrases like ``Let’s think step by step''~\citep{kojima2023large}. 
In addition to guiding the model through prompting methods, several works propose to control the reasoning path during inference through verifier~\citep{cobbe-etal-2020-training, khalifa2023discriminatorguided} and decoding method~\citep{obrien2023contrastive}. 
Another research line is a tailor-designed reasoning model that improves the mathematical reasoning ability of LLMs through instruction tuning on reasoning data, including reinforcement learning~\citep{uesato-etal-2022-solving, luo2023wizardmath}, knowledge distillation~\citep{fu2023specializing, hsieh-etal-2023-distilling, magister-etal-2023-teaching, shridhar-etal-2023-distilling, yue2024mammoth}, and data augmentation~\citep{huang-etal-2023-large, zelikman2022star, ni-etal-2023-learning, zhu-etal-2023-solving, yu2023metamath}. In this work we extend the multilingual reasoning carried out by \citet{shi2023language} in GPT-3 and PaLM~\citep{chowdhery2022palm} to current open-source popular LLMs, and study multilingual reasoning consistency. Additionally, following previous works~\citep{chen2023breaking, chai2024xcot}, we employ machine translation to translate existing English data into other languages for multilingual reasoning, but we do so at a very large scale, with substantial gains in performance. 

\paragraph{Consistency in Language Models} 
Consistency is one of the core qualities of language models, which refers to models behaving consistently on semantically equivalent inputs~\citep{elazar-etal-2021-measuring, fierro-sogaard-2022-factual}. Intra-language consistency has been studied across different tasks, such as language inference~\citep{li-etal-2019-logic, mitchell-etal-2022-enhancing}, explanation generation~\citep{camburu-etal-2020-make}, fill-in-the-blank phrases~\citep{ravichander-etal-2020-systematicity}, and math reasoning~\citep{wang2023selfconsistency}. These works mainly consider the English language, although there is some recent research on factual consistency in a multilingual scenario~\citep{fierro-sogaard-2022-factual, qi-etal-2023-cross}. To our knowledge, the present work is the first systematic analysis of multilingual reasoning consistency for LLMs, measuring the extent to which language models reason about the same answer to the same question written in different languages. In this context, we introduce multilingual CoT instruction tuning to boost reasoning capability across multiple languages, thereby improving model consistency.

\section{Multilingual Reasoning Consistency}
\label{sec:mrc}

This section provides details on the task, the data, and the experimental setup we use to investigate the LLMs' reasoning capability in different languages and their reasoning consistency.

\subsection{Task, Dataset, and Setup}

\paragraph{Task Definition} 
Generally, each math problem is fed to an LLM along with a set of manually written CoT exemplars, which is expected to generate all necessary intermediate steps up to and including the final answer. Given a set of math problems $\mathcal{M}$, each problem consists of a triplet $($\texttt{question}:$q_{x}$, \texttt{reasoning steps}:$r_{x}$, \texttt{final answer}:$a)$, where the problem and intermediate steps are written in a natural language $x \in \mathcal{L}$. We define the multilingual reasoning consistency of an LLM as the extent to which it reasons to the same answer for the same question asked in different languages, including correct consistency and incorrect consistency. For a given language pair $(x, y)$, correct consistency (CC) is the percentage of identical math problems written in that language pair for which the correct answer is predicted. Formally:

\begin{equation}\label{eq:consis}
    \textrm{CC}(x, y) = \frac{ {\textstyle \sum_{i=1}^{|M|}} \mathbb{I}\left(\hat{a}_{i}^{x}=\hat{a}_{i}^{y}=a_{i}\right)}{|M|}
\end{equation}

\noindent Where $\mathbb{I}(\cdot)$ is the indicator function. $a_{i}$ represents the gold answer corresponding to the $i$-{th} math question, $\hat{a}_{i}^{x}$ and $\hat{a}_{i}^{y}$ are the predicted answers to the $i$-{th} question written in languages $x$ and $y$,  respectively. 
For incorrect consistency (IC), we calculate the proportion of predicted answers that are incorrect but identical in the language pair out of the total number of incorrect answers in each respective language, and take the average of the two languages as the final result:

\begin{align}\label{eq:consis}
    \textrm{IC}(x, y) &= \left( \frac{ {\textstyle \sum_{i=1}^{|M|}} \mathbb{I}(\hat{a}_{i}^{x}=\hat{a}_{i}^{y} \neq a_{i})}{\sum_{i=1}^{|M|}\mathbb{I}(\hat{a}_{i}^{x} \neq a_{i})} \right. \notag \\ &\quad \left. + \frac{ {\textstyle \sum_{i=1}^{|M|}} \mathbb{I}(\hat{a}_{i}^{x}=\hat{a}_{i}^{y} \neq a_{i})}{\sum_{i=1}^{|M|}\mathbb{I}(\hat{a}_{i}^{y} \neq a_{i})} \right)/2
\end{align}

\paragraph{Reasoning Dataset} 
As defined above and depicted in Figure~\ref{fig:mcot}, multilingual reasoning consistency requires both questions and reasoning steps to be written in the same language, for multiple languages. GSM8K~\citep{cobbe-etal-2020-training} is an English (EN) dataset which includes high-quality grade school math word problems, each involving basic arithmetic operations (addition, subtraction, multiplication, and division) that usually require two to eight steps to solve according to the ofﬁcial, provided solution. It contains approximately 7,500 and 1,319 samples for training and testing, respectively. Based on GSM8K,~\citet{shi2023language} create MGSM, extending math reasoning into a multilingual setting. To do so, they select the first 250 problems from GSM8K and manually translate them into ten different languages: Bengali (BN), Chinese (ZH), French (FR), German (DE), Japanese (JA), Russian (RU), Spanish (ES), Swahili (SW), Telugu (TE) and Thai (TH). SW, BN, TE, and TH are usually heavily underrepresented languages in pre-trained language models; in PaLM~\citep{chowdhery2022palm} they account for less than 0.1\% of the pretraining data. We conduct experiments on these ten languages plus English, and measure the reasoning consistency between any two languages.

\paragraph{Model Setup}
We select a range of open-source state-of-the-art LLMs in three different sizes:


\begin{itemize}[leftmargin=*]
\itemsep 0in

\item 7B: LLAMA2~\citep{touvron2023llama}; Qwen~\citep{bai2023qwen}; Mistral~\citep{jiang2023mistral}.

\item 13-14B:  LLAMA2-13B~\citep{touvron2023llama}; Qwen-14B~\citep{bai2023qwen}.

\item 56-72B: Mistral-8$\times$7B\footnote{\url{https://mistral.ai/news/mixtral-of-experts/}}; LLAMA2-70B~\citep{touvron2023llama}; Qwen-72B~\citep{bai2023qwen}.

\end{itemize}

\noindent We perform few-shot CoT reasoning following \citet{wei2022chain} and \citet{shi2023language}, who improve the reasoning task by augmenting few-shot examples with intermediate steps. We use 8-shot for all languages except TE which only uses 2-shot due to the maximum number of input tokens. All CoT prompts in different languages are sourced from the original multilingual CoT reasoning paper~\citep{shi2023language}. On the other hand, assuming that we can not access existing math problems with the corresponding reasoning solutions in some languages, a natural and simple way is to use machine translation to translate existing data (e.g., English data) into the target languages. Therefore, we compare not only the consistency between languages, but also the consistency between human-translated (HT) prompts and machine-translated (MT) prompts to understand the impact of translated data on model performance.

\begin{figure*}[!t]
 \begin{minipage}[t]{0.7\linewidth}
    \centering
    \subfigure[Reasoning accuracy (\%) using human-written prompt. ]{
      \includegraphics[scale=0.63]{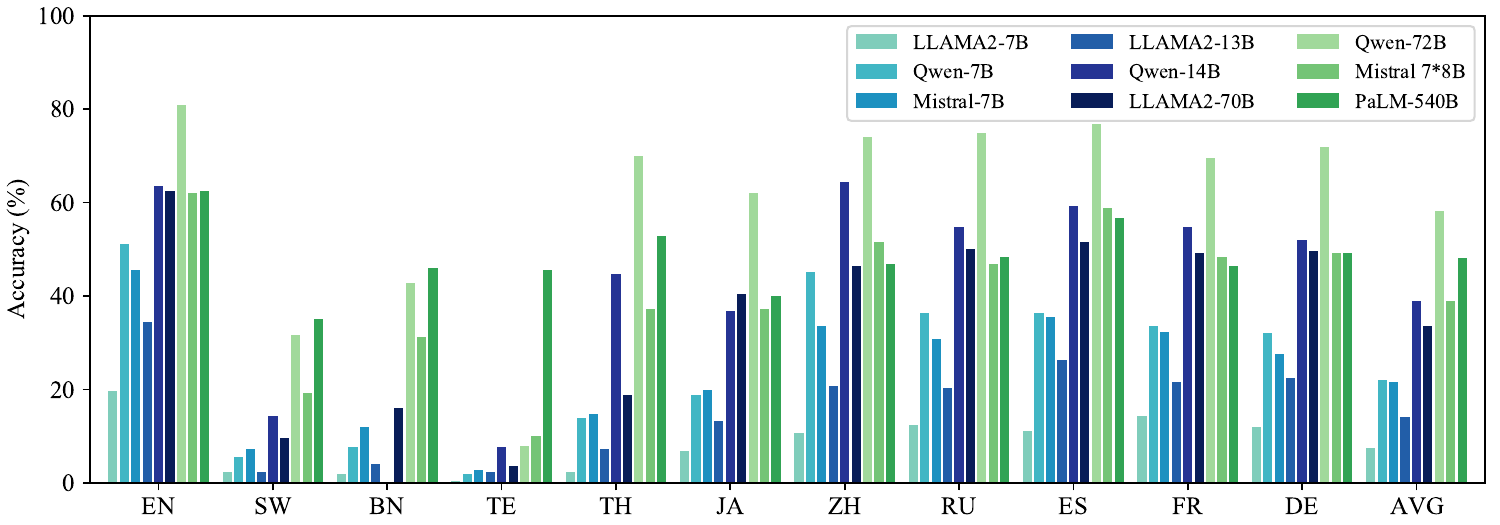}
      \label{fig:ht}
    }
    \end{minipage}

    \vspace{1.0mm}
    \begin{minipage}[t]{0.7\linewidth}
    \centering
    \subfigure[Reasoning accuracy (\%) using machine-translated prompt. ]{
      \includegraphics[scale=0.63]{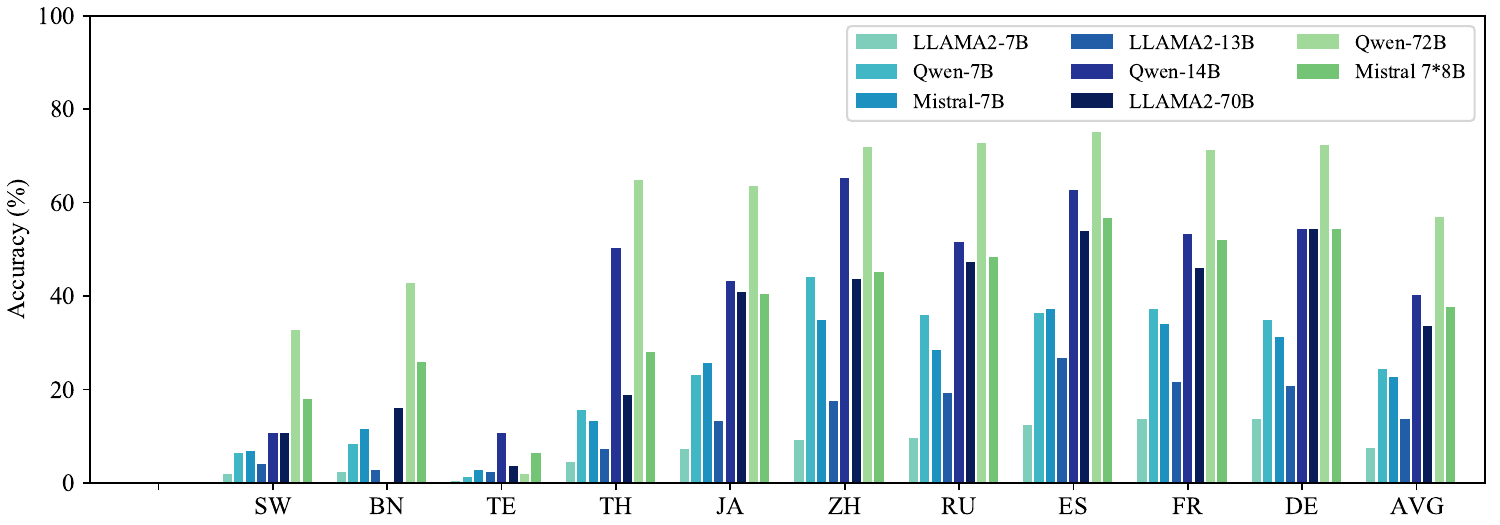}
      \label{fig:mt}
    }
    \end{minipage}
\caption{Accuracy (\%) on MGSM of different models with the few-shot method. All machine-translated prompts are translated from English data using Google Translate.} 
\label{fig:acc-mgsm}
\end{figure*}

\begin{figure*}[!t]
 \begin{minipage}[t]{0.7\linewidth}
    \centering
    \subfigure[Reasoning consistency (\%) between language pairs in LLMs. ]{
      \includegraphics[scale=0.79]{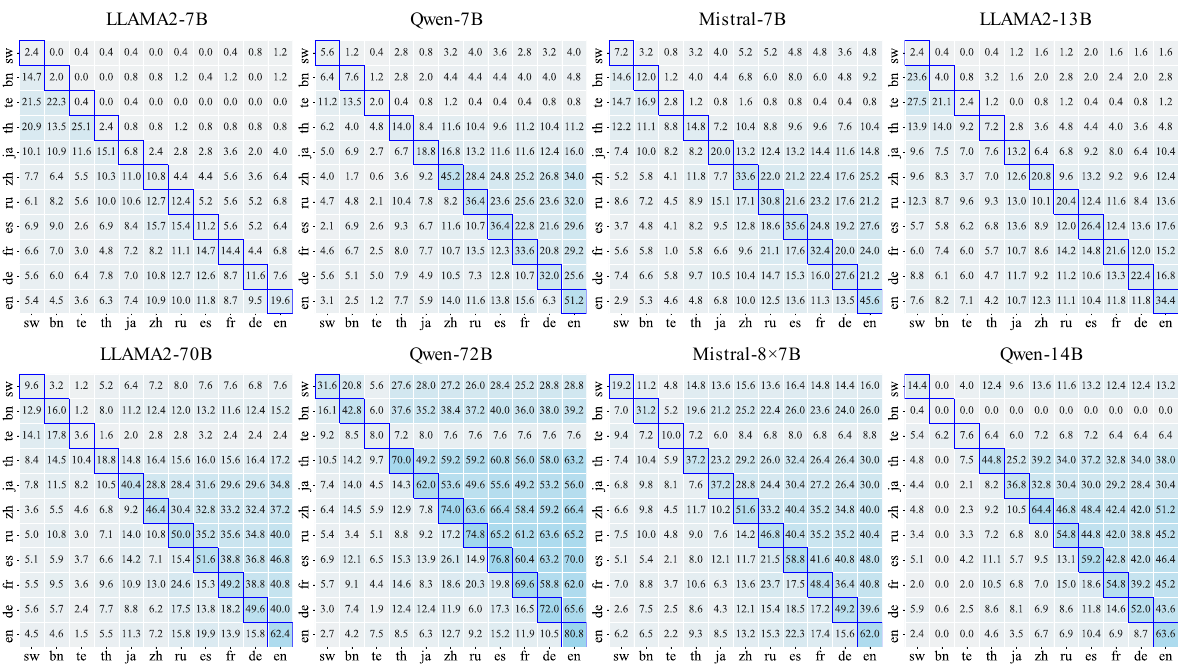}
      \label{fig:inl}
    }
    \end{minipage}

    \vspace{1.0mm}
    \begin{minipage}[t]{0.7\linewidth}
    \centering
    \subfigure[Reasoning consistency (\%) between human-translated (row) and machine-translated (column) prompts in LLMs. ]{
      \includegraphics[scale=0.79]{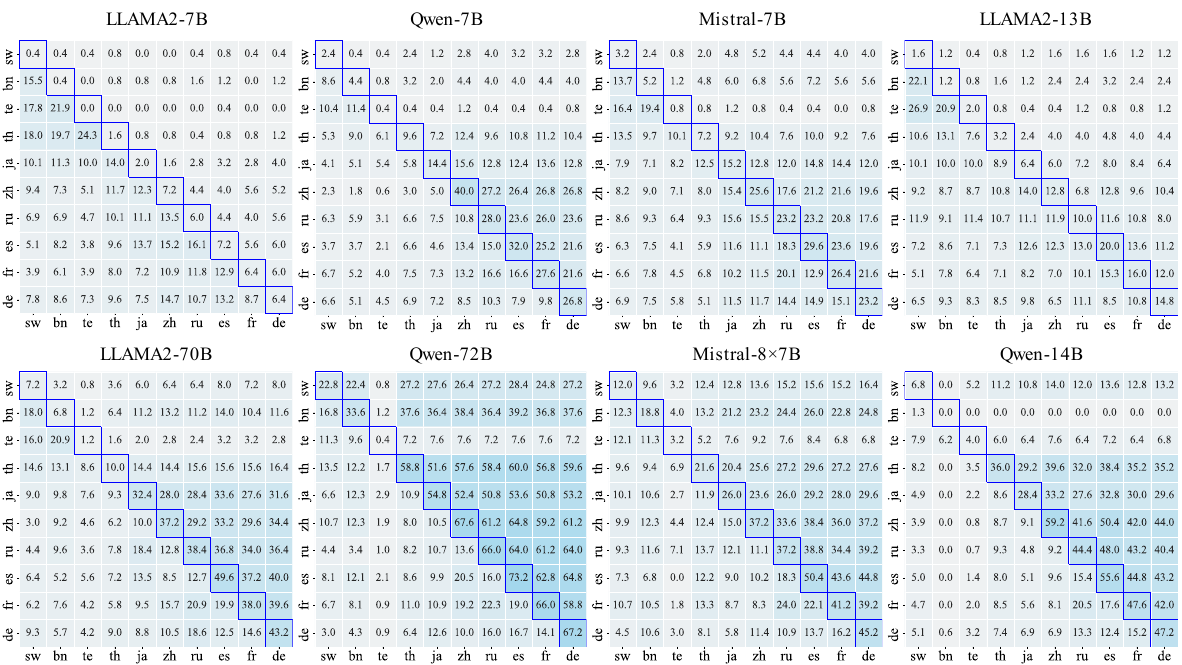}
      \label{fig:mht}
    }
    \end{minipage}
\caption{Multilingual reasoning consistency. The triangle above the marked diagonal shows the consistency of the models on the correct answers; the triangle below the diagonal contains the consistency between the language pairs where the final answer is the same but incorrect.} 
\label{fig:mrc}
\end{figure*}

\subsection{Results}

\paragraph{Model Performance}
We first compare the performance of different models, as shown in Figure~\ref{fig:acc-mgsm}.\footnote{Detailed results are in Appendix~\ref{app:mgsm}.} Unsurprisingly, and similar to the results reported by~\citet{shi2023language}, the performance generally improves for all models across all languages as the models scale up in size. However, in contrast to their findings for the large-scale model PaLM-540B, open-source models still yield a performance gap between underrepresented languages (which cover less than 0.1\% of the training corpora in PaLM) and high-resource languages. For instance, Qwen-7B scores above 30\% in most high-resource languages (e.g., ES and FR) and below 10\% in underrepresented languages (e.g., SE, BN, and TE). Particularly, Mistral-8$\times$7B and Qwen-72B achieve similar or higher scores than PaLM-540B in high-resource languages, while PaLM-540B has better results in underrepresented languages. Regarding HT and MT prompts, we compare all languages except EN, and observe that the results using these two prompts are very close for all models and languages we considered.

\paragraph{Reasoning Consistency}
We illustrate multilingual reasoning consistency results for different models in Figure~\ref{fig:mrc}. 
When looking at reasoning consistency between language pairs as presented in Figure~\ref{fig:inl}, consistent with the trend in model performance, we find overall consistency to be lower for smaller-scale models (7B and 13B) and underrepresented languages (SW, BN, and TE). The correct consistency for all models improves with increasing language representation in pre-training, which is not surprising as higher-resource languages have better accuracy. Incorrect consistency, however, shows a different trend, with higher scores between underrepresented languages and between high-resource languages in most models, and this occurs even in larger-scale models such as LLAMA2-70B and Qwen-72B. This suggests that incorrect reasoning knowledge in LLMs is similar to a certain extent between these languages.

To further analyse the impact of using machine-translated prompts, we present reasoning consistency between HT and MT prompts in Figure~\ref{fig:mht}. The consistency trend is similar to those in Figure~\ref{fig:inl}, with high-resource languages having high consistency on correct answers and underrepresented languages having high consistency on incorrect answers. On the other hand, consistency scores of the larger models are closer to the reasoning accuracy of using human-written prompts, indicating that using machine-translated prompts has less impact on these models.

\begin{figure}[!t]
\centering
\includegraphics[scale=.55]{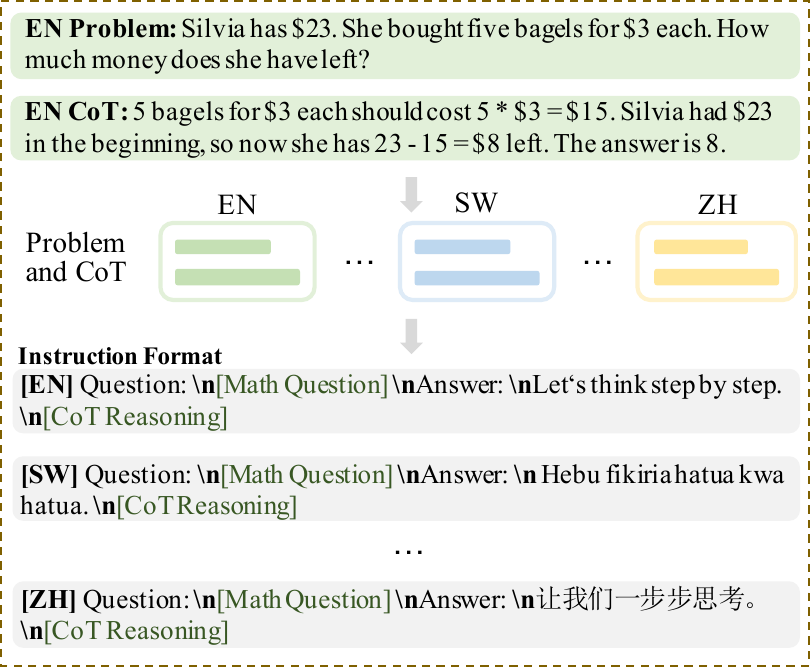}
\caption{Overview of multilingual CoT reasoning data. English data is first automatically translated into target languages, and then inserted into the templates to construct multilingual instruction data.} 
\label{fig:data}
\end{figure}

\section{mCoT Instruction Tuning}
Based on our findings in multilingual reasoning consistency, we propose a multilingual instruction tuning framework mCoT to supervise the reasoning process, aiming to generate similar reasoning and the same results on inputs expressed in different languages but semantically equivalent. 
Formally, given a question $q=\{q_{1}, \cdots, q_{m}\}$ in language $x$ and its corresponding answer -- always an integer in this work -- the model can generate the step-by-step reasoning solution $s=\{s_{1}, \cdots, s_{n}\}$ written in language $x$ with the final answer.

\subsection{mCoT-MATH}
\paragraph{Source Data}
We leverage two English datasets as source data: MetaMathQA~\citep{yu2023metamath}, which augments the training data of GSM8K and MATH with a question bootstrapping method that rewrites questions using both forward and backward reasoning paths and leverages LLMs to reformulate the question text; and 
MathInstruct~\citep{yue2024mammoth}, which is based on seven existing math rationale datasets annotated by humans or GPT-4~\citep{openai2023gpt4}. This dataset contains two prompt formats: Chain-of-Thought (CoT, \citealt{wei-2022-chain}) and Program-of-Thought (PoT, \citealt{chen2023program}). We select MathInstruct's CoT samples and combine them with MetaMathQA, resulting in approximately 580,000 samples. 

\paragraph{Automatic Translation}
Following~\citet{shi2023language}, we select the ten different languages included in MGSM as the target languages for translation. The overall framework is illustrated in Figure~\ref{fig:data}. First, we machine-translate\footnote{\url{https://translate.google.com/}.} all EN data into the target languages. After translation, we use the instruction format to reformulate all the data to obtain mCoT-MATH, the first large-scale multilingual math CoT dataset, containing around 6.3 million samples. This is expected to facilitate an exhaustive exploration of model reasoning consistency across languages.

\subsection{Implementation}
We use Mistral-7B as base model, training our mCoT instruction tuning framework 
using HuggingFace Transformers~\citep{wolf-etal-2020-transformers} and Deepspeed~\citep{rasley2020deepspeed}. 
During training, we set the maximum length of the input sequence to 1024, thus reducing GPU memory consumption and improving training speed. We train our model using AdamW optimiser~\citep{loshchilov2018decoupled} with a maximum learning rate of 5e-6 and a 3\% learning rate warmup. The batch size is set to 32 and gradients are accumulated in 4 update steps. 
We train our model on 4 $\times$ NVIDIA A100 40GB GPUs for around 10 days. 
We report the final answer accuracy for all experiments.

\subsection{Evaluation Data} We select two popular multilingual math reasoning datasets, in which each sample contains a question and the corresponding final answer. Specifically, in addition to MGSM, we also include MSVAMP~\citep{chen2023breaking}, which is constructed based on English data SVAMP~\citep{patel-etal-2021-nlp}. \citet{chen2023breaking} use Google Translate to transform 1,000 questions from the SVAMP test set into nine languages: SW, BN, TH, JA, ZH, RU, ES, FR, and DE. To ensure the translation quality, they back-translate the translated text into English and ask three professional annotators to check semantic consistency manually.

\subsection{Baselines}
 We consider state-of-the-art models of different sizes, both closed and open source.

\paragraph{Close-Source Models} We include six models from three different companies as foundational benchmarks: (i) OpenAI's GPT-3~\citep{brown-etal-2020-language}, GPT-3.5\footnote{\url{https://openai.com/blog/chatgpt}.}, and GPT-4~\citep{openai2023gpt4}; (ii) Anthropic’s Claude-2\footnote{\url{https://www.anthropic.com/index/claude-2}.}; and (iii) Google’s PaLM 2~\citep{chowdhery2022palm} and Flan-PaLM~\citep{anil2023palm}.

\paragraph{Open-Source Models}
We also compare our model mCoT with several best-performing open-source models for the sake of fairness (including size comparison): 
(i) WizardMath~\citep{luo2023wizardmath}; (ii) MathOctopus~\citep{chen2023breaking}; (iii) xCoT~\citep{chai2024xcot}; and (iv) MetaMath~\citep{yu2023metamath}.

\begin{table*}[!t]
\centering
\setlength{\tabcolsep}{6pt}
\resizebox{\linewidth}{!}{%
\begin{tabular}{lrrrrrrrrrrr}
\toprule
\textbf{Model} &  \makecell[c]{\textbf{SW}} & \makecell[c]{\textbf{BN}} & \makecell[c]{\textbf{TE}} & \makecell[c]{\textbf{TH}} & \makecell[c]{\textbf{JA}} & \makecell[c]{\textbf{ZH}} & \makecell[c]{\textbf{RU}} & \makecell[c]{\textbf{ES}} & \makecell[c]{\textbf{FR}} & \makecell[c]{\textbf{DE}} & \makecell[c]{\textbf{EN}}\\
\hline
Lang. Freq. (\%) & <0.1 & <0.1 & <0.1 & <0.1 & 0.4 & 0.4 & 0.5 & 2.1 & 3.3 & 3.5 & 78.0 \\
\hline
\multicolumn{12}{l}{\textbf{Close-Source Models }}\\
\hline
GPT-3 few-shot     & 11.2 & 6.4 & 0.4 & 0.8 & 26.0 & 40.0 & 28.4 & 40.4 & 37.6 & 36.0 & 53.6\\
GPT-3.5-En 2-shot   & 40.0 & 7.6 & \makecell[c]{-} & 15.6 & 46.8 & 52.8 & 50.4 & 61.2 & 59.2 & 62.0 & 67.2\\
GPT-4-En 2-shot   & \textbf{64.4} & 17.6 & \makecell[c]{-} & 40.4 & \textbf{71.6} & \textbf{70.0} & \textbf{64.0} & \textbf{71.2} & \textbf{72.0} & \textbf{73.6} & \textbf{80.0} \\
PaLM-540B few-shot & 35.2 & \textbf{46.0} & \textbf{45.6} & \textbf{52.8} & 40.0 & 46.8 & 48.4 & 56.8 & 46.4 & 49.2 & 62.4\\
\midrule
\textbf{Open-source Models}\\
\hline
7B Models\\
WizardMath~\citep{luo2023wizardmath} & 3.4 & 2.0 & \makecell[c]{-} & 4.0 & 24.0 & 22.4 & 30.8 & 34.8 & 30.4 & 30.4 & 47.6 \\
MathOctopus~\citep{chen2023breaking} & 38.4 & 33.2 & \makecell[c]{-} & 36.4 & 35.6 & 45.2 & 48.4 & 45.2 & 38.0 & 43.6 & 54.8 \\
MathOctopus-Mistral~\citep{chen2023breaking} & 51.6 & 44.0 & \makecell[c]{-} & 48.8 & 48.0 & 51.6 & 49.6 & 53.2 & 47.2 & 50.0 & 58.4 \\
xCoT~\citep{chai2024xcot}    & 48.4 & 40.4 & 42.8 & 49.2 & 50.0 & 50.0 & 50.0 & 48.8 & 49.6 & 47.2 & 48.4\\
\arrayrulecolor{black}\hline
13B Models\\
WizardMath~\citep{luo2023wizardmath} & 5.6 & 6.4 & \makecell[c]{-} & 5.6 & 22.0 & 28.0 & 34.4 & 45.6 & 42.0 & 40.4 & 52.8 \\
MathOctopus~\citep{chen2023breaking} & 46.0 & 42.0 & \makecell[c]{-} & 46.0 & 39.6 & 51.2 & 47.6 & 53.2 & 49.6 & 49.2 & 51.6\\
xCoT~\citep{chai2024xcot} & 51.6 & 50.0 & 47.2 & 50.0 & 49.6 & 54.0 & 56.8 & 54.8 & 46.4 & 52.4 & 54.4 \\
\arrayrulecolor{black}\hline
mCoT-7B (ours) & \textbf{67.2} & \textbf{65.6} & \textbf{62.4} & \textbf{67.6} & \textbf{65.2} & \textbf{64.8} & \textbf{66.8} & \textbf{68.4} & \textbf{63.8} & \textbf{61.2} & \textbf{71.6}\\
\bottomrule
\end{tabular}}
\caption{\label{tab:acc-main}
Multilingual evaluation results (final answer accuracy:\%) on the MGSM benchmark. Notes: (i) Lang. Freq. (\%) is the language frequency in PaLM training data; (ii) the results of GPT-3 and PaLM-540B are from~\citet{shi2023language}, while those for GPT-3.5 and GPT-4 are from~\citet{chen2023breaking}; and (iii) in boldface best results per language among closed models and among open models.
}
\end{table*}

\begin{table*}[!t]
\centering
\setlength{\tabcolsep}{6pt}
\resizebox{\linewidth}{!}{%
\begin{tabular}{lrrrrrrrrrrr}
\toprule
\textbf{Model} &  \makecell[c]{\textbf{SW}} & \makecell[c]{\textbf{BN}} & \makecell[c]{\textbf{TH}} & \makecell[c]{\textbf{JA}} & \makecell[c]{\textbf{ZH}} & \makecell[c]{\textbf{RU}} & \makecell[c]{\textbf{ES}} & \makecell[c]{\textbf{FR}} & \makecell[c]{\textbf{DE}} & \makecell[c]{\textbf{EN}} & \makecell[c]{\textbf{AVG}} \\
\hline
Lang. Freq. (\%) & <0.1 & <0.1 & <0.1 & 0.4 & 0.4 & 0.5 & 2.1 & 3.3 & 3.5 & 78.0 & \makecell[c]{-}\\
\hline
\multicolumn{12}{l}{\textbf{Close-Source Models}}\\
\hline
GPT-3.5-En zero-shot  & 63.2 & 3.1 & 24.4 & 63.3 & 72.4 & 62.3 & 69.5 & 71.9 & 66.7 & 76.1 & 57.3\\
GPT-3.5-En 2-shot  & 68.4 & 14.4 & 46.0 & 74.0 & 78.4 & 70.9 & 74.6 & 78.2 & 73.9 & 81.2 & 66.0\\
GPT-4-En 2-shot     & \textbf{75.7} & \textbf{31.2} & \textbf{68.1} & \textbf{74.8} & \textbf{78.9} & \textbf{77.9} & \textbf{81.5} & \textbf{83.9} & \textbf{78.1} & \textbf{80.1} & \textbf{73.0}\\
\midrule
\textbf{Open-source Models}\\
\hline
7B Models\\
WizardMath~\citep{luo2023wizardmath} & 10.3 & 16.1 & 6.3 & 26.7 & 26.8 & 33.7 & 42.9 & 39.9 & 39.6 & 45.1 & 27.0\\
MathOctopus~\citep{chen2023breaking} & 42.3 & 32.8 & 40.5 & 43.2 & 43.2 & 42.1 & 44.5 & 45.3 & 43.1 & 46.8 & 42.4\\
MathOctopus-Mistral~\citep{chen2023breaking} & 41.2 & 36.7 & 40.2 & 41.5 & 43.1 & 44.0 & 47.0 & 49.0 & 46.4 & 49.7 & 43.9\\
\arrayrulecolor{black}\hline
13B Models\\
WizardMath~\citep{luo2023wizardmath} & 12.5 & 13.7 & 16.3 & 29.5 & 37.0 & 43.8 & 50.4 & 49.4 & 48.7 & 56.3 & 35.8\\
MathOctopus~\citep{chen2023breaking} & 43.4 & 34.2 & 39.5 & 43.1 & 46.4 & 48.2 & 48.2 & 49.9 & 47.7 & 44.6 & 44.5\\
\arrayrulecolor{black}\hline
mCoT-7B (ours) & \textbf{55.0} & \textbf{53.7} & \textbf{56.4} & \textbf{58.8} & \textbf{58.2} & \textbf{58.1} & \textbf{58.9} & \textbf{58.8} & \textbf{61.1} & \textbf{58.3} & \textbf{57.7}\\
\bottomrule
\end{tabular}}
\caption{\label{tab:acc-msvamp}
Multilingual evaluation results (final answer accuracy:\%) on the MSVAMP benchmark. Notes: (i) Lang. Freq. (\%) is the language frequency in PaLM training data; (ii) the results of GPT-3.5 and GPT-4 are from~\citet{chen2023breaking}; and (iii) in boldface best results per language among closed models and among open models.
}
\end{table*}

\subsection{Results}

\paragraph{Evaluation on MGSM}
Table~\ref{tab:acc-main} reports results on the MGSM benchmark. The first observation is that most existing models, including close- and open-source, perform poorly on underrepresented languages such as SW, BN, and TE. For close-source models, GPT models achieve higher accuracy in high-resource languages, with GPT-4 scoring the highest; PaLM-540 achieves competitive performance in all languages, especially reaching the highest score in low-resource languages BN, TE and TH. For open-source models, we observe that WizardMath achieves less than 7\% accuracy on low-resource languages which is explained by the fact that this model is trained on English data, while both MathOctopus and xCoT gain strong improvement with the help of the multilingual instruction dataset. When looking at our model, we see that mCoT significantly outperforms all previous strong baselines, and even outperforms GPT-4 in underrepresented languages such as SW, BN, TE, and TH. Particularly, mCoT has higher accuracy scores than PaLM-540B across all languages.

\paragraph{Evaluation on MSVAMP}
Table~\ref{tab:acc-msvamp} reports results on the  MSVAMP benchmark. We can observe that GPT-4 with 2-shot achieves the best performance in all languages except BN, where our model achieves the best results. When looking only at open-source models, similar to the observations on MGSM, WizardMath performs poorly in low-resource languages, while our model mCoT shows the highest scores across the board. In particular, mCoT outperforms MathOctopus-Mistral, the one also based on Mistral-7B, confirming that leveraging our dataset mCoT-MATH can yield substantial gains in
performance. Finally, we observe that mCoT scores are very close across all languages, suggesting a lesser dependency on the low- vs high-resource aspect.

\begin{figure}[!t]
    \centering
    \includegraphics[scale=.36]{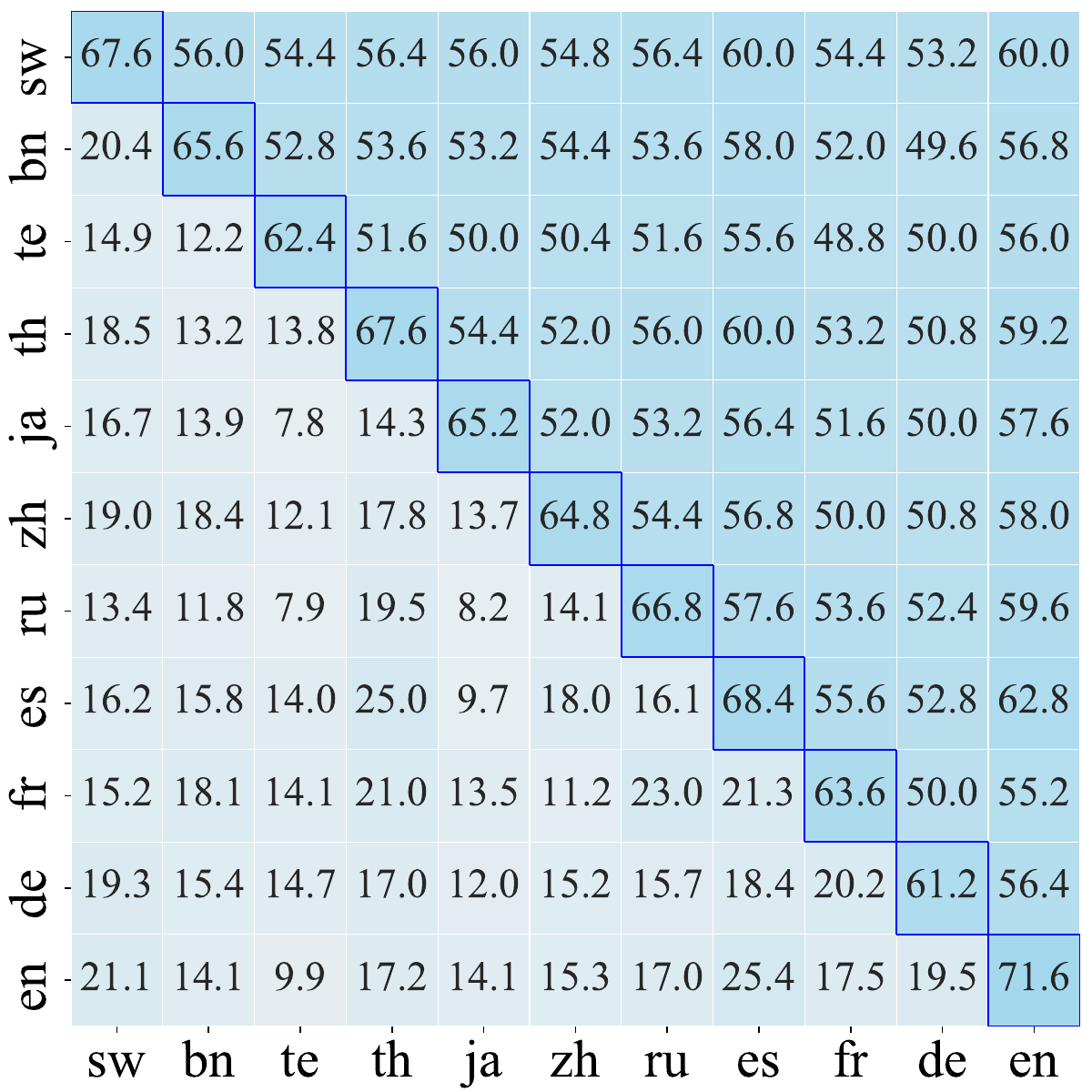}
\caption{Multilingual reasoning consistency of mCoT. The triangle above the marked diagonal shows the consistency of the models on the correct answers; the triangle below the diagonal contains the consistency between the language pairs where the final answer is the same but incorrect.} 
\label{fig:mcot-consistency}
\end{figure}

\paragraph{Reasoning Consistency} 
To further evaluate mCoT's reasoning capability in different languages, in Figure~\ref{fig:mcot-consistency} we present its reasoning consistency results on the dataset MGSM. After instruction tuning on our dataset mCoT-MATH, correct consistency shows a strong improvement as our model's reasoning accuracy improves, and in particular we observe that: (i) this is especially true for underrepresented languages; and (ii) the scores for all language pairs are very close, with most being above 50\%. It is also interesting to see an increasing trend in incorrect consistency for most language pairs, including low- and high-resource languages, confirming that our model exhibits similar reasoning capabilities across languages. Overall, these observations underscore the efficacy of our method in improving the model's reasoning consistency.

\begin{table*}[t]
    \centering
    \begin{tabular}{c}
    \includegraphics[width=0.99\textwidth]{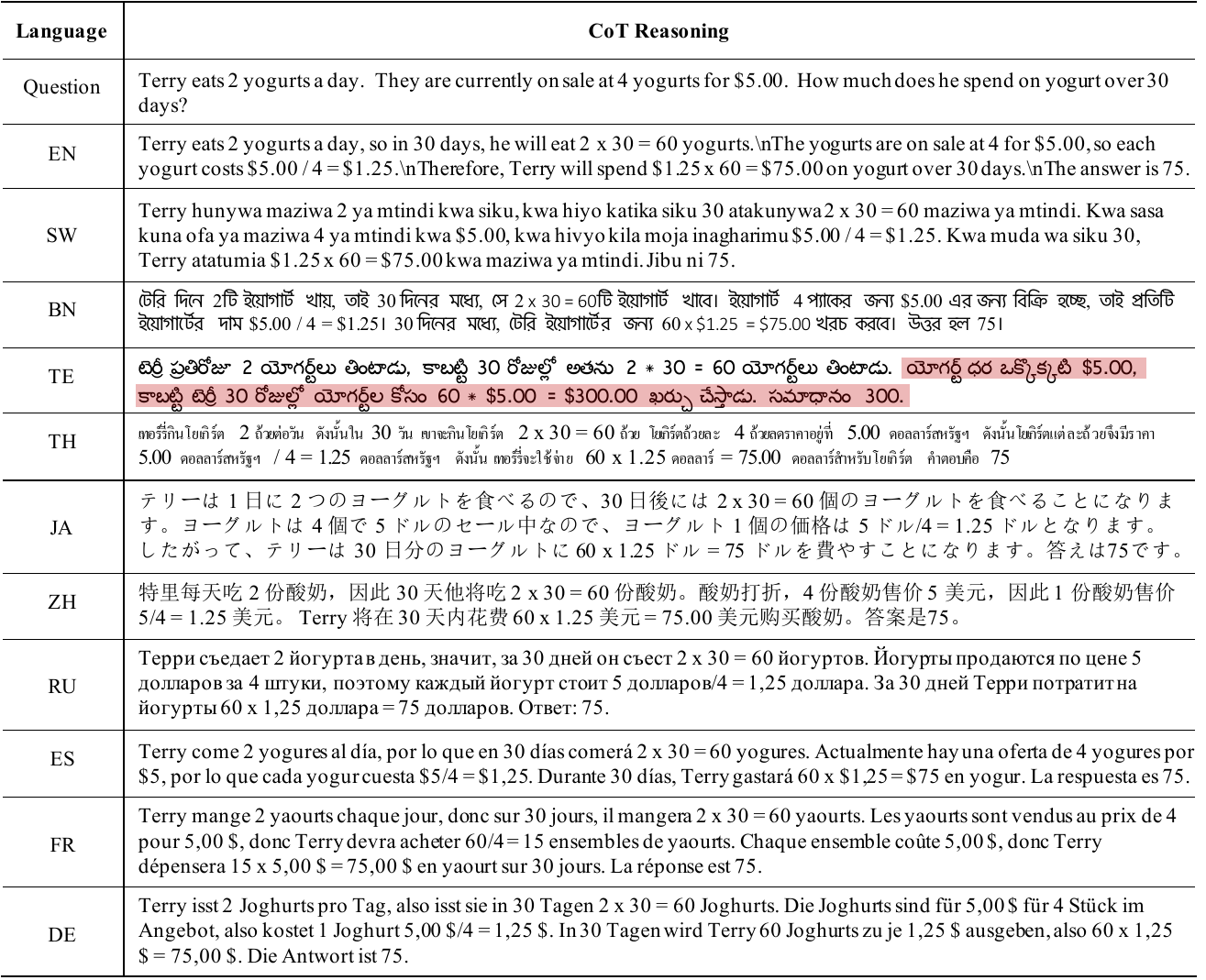}
    \end{tabular}
    \caption{Case study in the test set of MSGM. Note that here we only show an EN question, the questions corresponding to each output are written in their respective languages.}
    \label{tab:example}
\end{table*}

\section{Analysis}

\paragraph{Case Study}
Table~\ref{tab:example} shows reasoning solutions generated by our model mCoT for the same math question expressed in different languages. In this case, mCoT incorrectly reasons in the second step in TE: \textit{Yogurts cost \$5.00 each, so Terry spends 60 * $5.00 = $300.00 on yogurts in 30 days} (EN translation of red background part), which leads to the wrong reasoning and final answer. For other languages, we observe that mCoT generates the correct final answers, while the solutions might be logically different. For example, the focus of FR is to first reason about how many packs of yogurt are needed and then calculate the total cost based on the price of each pack, while for other languages such as EN, the total cost is calculated through the total amount of yogurt and the unit price of each yogurt. Overall, these evidences demonstrate that mCoT has a good reasoning capability across different languages.

\section{Conclusion}
 We studied multilingual reasoning consistency across multiple languages for popular open-source LLMs such as Mistral and LLAMA2, which provides insights into the evaluation of LLMs. Our findings show that there is a substantial variation across the languages, with lesser resourced ones substantially underperforming. To address this issue, we constructed the first large-scale multilingual math reasoning instruction dataset mCoT-MATH, with around 6.3 million samples in eleven diverse languages. We then introduced a multilingual reasoning instruction tuning framework to train our model mCoT on mCoT-MATH. Evaluation on two multilingual benchmark datasets shows that our 7B parameter model achieves impressive reasoning consistency across all languages, and comparable or superior performance to close- and open-source state-of-the-art models even of much larger sizes.

\section{Limitations}
In this work we investigated multilingual reasoning consistency across 11 languages and eight open-source models in different sizes, but there are very many more languages and LLMs that can still bring substantial challenges and insights if considered. In addition, when considering more languages in the future, it will also be interesting to consider language families as factor. On the other hand, this work focused on reasoning consistency based on the final answer, while the consistency of intermediate reasoning steps is definitely an interesting direction. Reasoning solutions are not necessarily consistent across languages (even within the same language), since they might be logically different but still result in the same and correct answer. Therefore, automatically assessing intermediate steps is challenging and requires more explorations in the future. Regarding the dataset, while mCoT-MATH boosts the reasoning capability across languages, screening high-quality machine-translated data could further improve the model. Finally, the full potential of our approach could be further explored by for example extending instruction tuning with reward learning to encourage models to generate more diverse solutions.

\section*{Acknowledgments}

The anonymous reviewers of ACL Rolling Review provided us with useful comments which contributed to improving this paper and its presentation, so we’re grateful to them. We would also like to thank the SURF organisation and the Center for Information Technology of the University of Groningen for their support and for providing access to the high-performance computing clusters Snellius and Hábrók, respectively.

\bibliography{anthology,custom}
\bibliographystyle{acl_natbib}

\clearpage
\appendix
\onecolumn
\section{Appendix}

\subsection{Prompt Examples}

\begin{figure*}[h]
    \centering
    \subfigure[English CoT prompt.]{
      \includegraphics[scale=0.425]{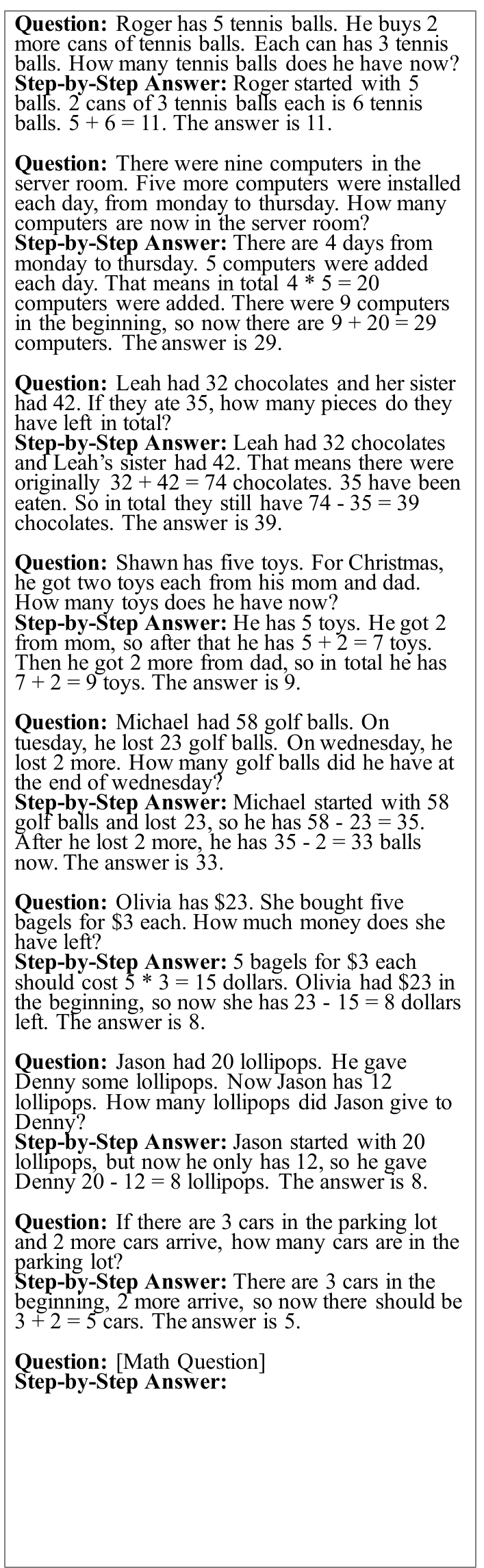}
      \label{fig:prompt-en}
    }
    \subfigure[HT German CoT prompt.]{
      \includegraphics[scale=0.425]{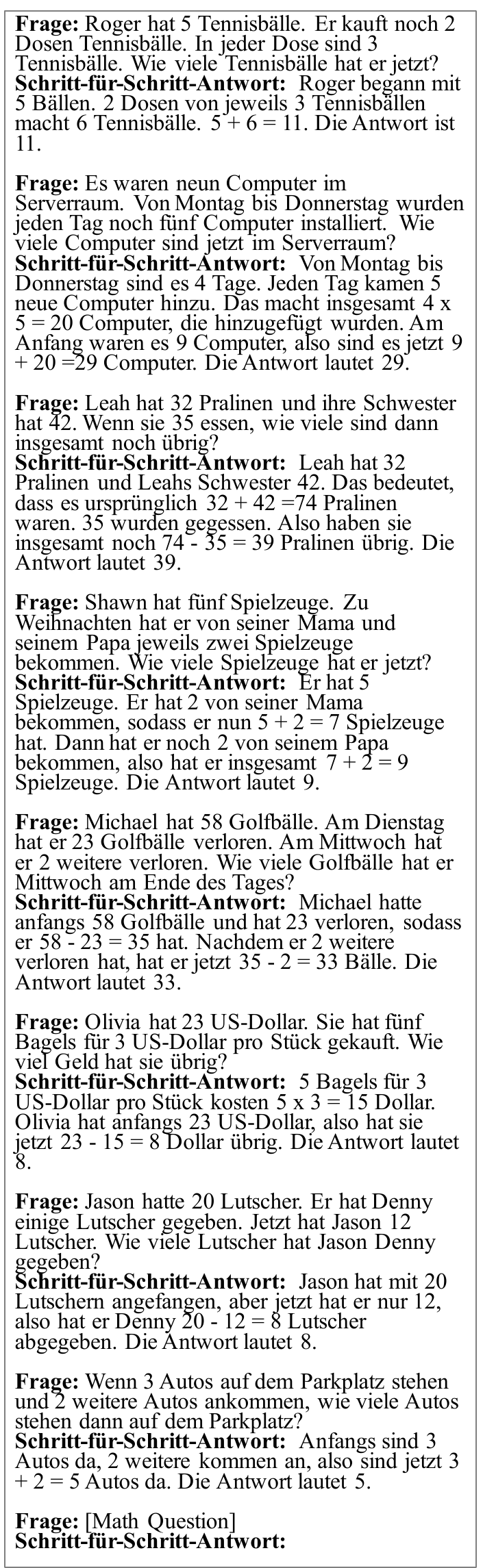}
      \label{fig:prompt-ht}
    }
    \subfigure[MT German CoT prompt.]{
      \includegraphics[scale=0.425]{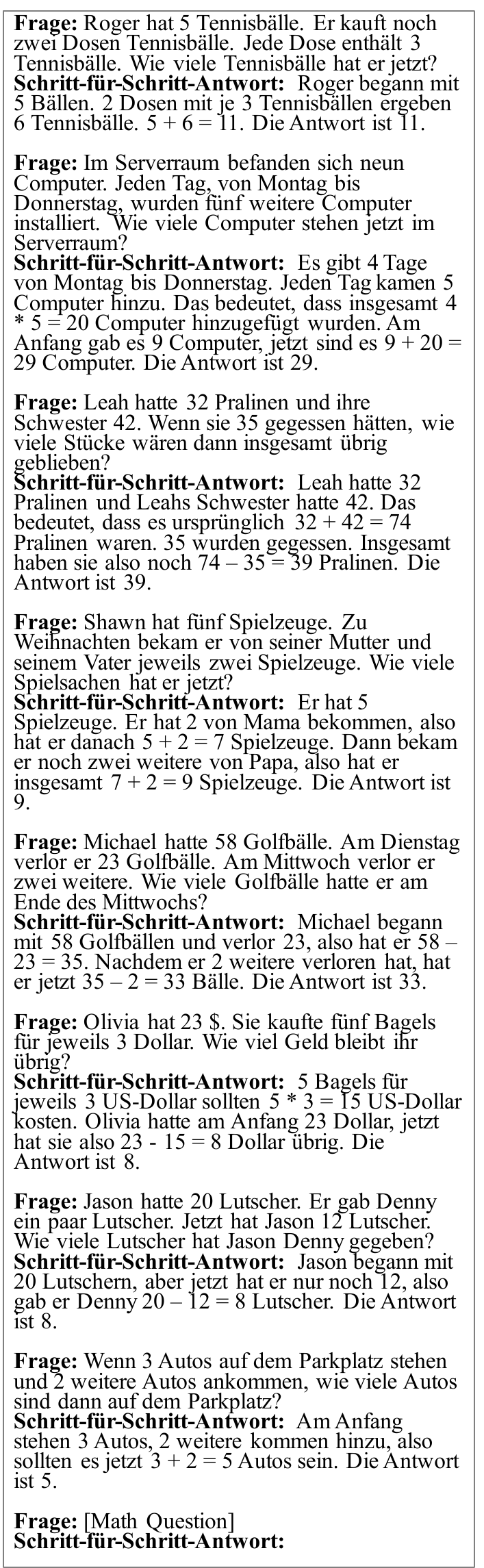}
      \label{fig:prompt-mt}
    }
\caption{CoT prompt template: mathematical questions are inserted in square brackets, and the model generates corresponding CoT reasoning.} 
\label{fig:samples}
\end{figure*}



\newpage
\subsection{Multilingual reasoning results on MGSM}
\label{app:mgsm}

\begin{table*}[h]
\centering
\setlength{\tabcolsep}{7pt}
\resizebox{\linewidth}{!}{%
\begin{tabular}{lcrrrrrrrrrrrr}
\toprule
\textbf{Language Model} & \textbf{Prompt} & {\textbf{EN}} & \makecell[c]{\textbf{SW}} & \makecell[c]{\textbf{BN}} & \makecell[c]{\textbf{TE}} & \makecell[c]{\textbf{TH}} & \makecell[c]{\textbf{JA}} & \makecell[c]{\textbf{ZH}} & \makecell[c]{\textbf{RU}} & \makecell[c]{\textbf{ES}} & \makecell[c]{\textbf{FR}} & \makecell[c]{\textbf{DE}} & \makecell[c]{\textbf{AVG}} \\
\hline
Lang. Freq. (\%) & - & 78.0 & <0.1 & <0.1 & <0.1 & <0.1 & 0.4 & 0.4 & 0.5 & 2.1 & 3.3 & 3.5 & \makecell[c]{-}\\
COMET Score & - & \makecell[c]{-} & 84.6 & 87.3 & 89.7 & 81.3 & 86.3 & 89.2 & 86.3 & 87.9 & 88.5 & 88.8 & \makecell[c]{-}\\
\hline
7B Models\\
\arrayrulecolor{black}\hline
\multirow{2}{*}{LLAMA2} & HT & 19.6 & 2.4 & 2.0 & 0.4 & 2.4 & 6.8 & 10.8 & 12.4 & 11.2 & 14.4 & 11.9 & 7.5\\
                         & MT & \makecell[c]{-} & 2.0 & 2.4 & 0.4 & 4.4 & 7.2 & 9.2 & 9.6 & 12.4 & 13.6 & 13.6 & 7.5\\
\multirow{2}{*}{Qwen}    & HT & 51.2 & 5.6 & 7.6 & 2.0 & 14.0 & 18.8 & 45.2 & 36.4 & 36.4 & 33.6 & 32.0 & 22.1\\
                         & MT & \makecell[c]{-} & 6.4 & 8.4 & 1.2 & 15.6 & 23.2 & 44.0 & 36.0 & 36.4 & 37.2 & 34.8 & 24.3\\
\multirow{2}{*}{Mistral} & HT & 45.6 & 7.2 & 12.0 & 2.8 & 14.8 & 20.0 & 33.6 & 30.8 & 35.6 & 32.4 & 27.6 & 21.7\\
                         & MT & \makecell[c]{-} & 6.8 & 11.6 & 2.8 & 13.2 & 25.6 & 34.8 & 28.4 & 37.2 & 34.0 & 31.2 & 22.6\\
\midrule
13-14B Models\\
\arrayrulecolor{black}\hline
\multirow{2}{*}{LLAMA2-13B} & HT & 34.4 & 2.4 & 4.0 & 2.4 & 7.2 & 13.2 & 20.8 & 20.4 & 26.4 & 21.6 & 22.4 & 14.1\\
                         & MT & \makecell[c]{-} & 4.0 & 2.8 & 2.4 & 7.2 & 13.2 & 17.6 & 19.2 & 26.8 & 21.6 & 20.8 & 13.6\\
\multirow{2}{*}{Qwen-14B} & HT & 63.6 & 14.4 & 0.0 & 7.6 & 44.8 & 36.8 & 64.4 & 54.8 & 59.2 & 54.8 & 52.0 & 38.9 \\
                      & MT & \makecell[c]{-} & 10.8 & 0.0 & 10.8 & 50.4 & 43.2 & 65.2 & 51.6 & 62.8 & 53.2 & 54.4 & 40.2\\
\midrule
$>$65B Models\\
\arrayrulecolor{black}\hline
LLAMA2-70B             & HT & 62.4 & 9.6 & 16.0 & 3.6 & 18.8 & 40.4 & 46.4 & 50.0 & 51.6 & 49.2 & 49.6 & 33.5\\
                      & MT & \makecell[c]{-} & 10.8 & 16.0 & 3.6 & 18.8 & 40.8 & 43.6 & 47.2 & 54.0 & 46.0 & 54.4 & 33.5\\
Qwen-72B              & HT & 80.8 & 31.6 & 42.8 & 8.0 & 70.0 & 62.0 & 74.0 & 74.8 & 76.8 & 69.6 & 72.0 & 58.2\\
                    & MT & \makecell[c]{-} & 32.8 & 42.8 & 2.0 & 64.8 & 63.6 & 72.0 & 72.8 & 75.2 & 71.2 & 72.4 & 57.0\\
Mistral-8$\times$7B   & HT & 62.0 & 19.2 & 31.2 & 10.0 & 37.2 & 37.2 & 51.6 & 46.8 & 58.8 & 48.4 & 49.2 & 39.0\\
                      & MT & \makecell[c]{-} & 18.0 & 26.0 & 6.4 & 28.0 & 40.4 & 45.2 & 48.4 & 56.8 & 52.0 & 54.4 & 37.6\\
PaLM-540B$^\dag$ & HT & 62.4 & 35.2 & 46.0 & 45.6 & 52.8 & 40.0 & 46.8 & 48.4 & 56.8 & 46.4 & 49.2 & 48.1\\
\bottomrule
\end{tabular}}
\caption{\label{tab:acc-few-shot}
Accuracy (\%) on MGSM of different models with the few-shot method. Notes: (i) Lang. Freq. (\%) is the language frequency in PaLM training data; (ii) we report COMET~\citep{rei-etal-2020-comet} score between the HT and MT prompts; (iii) average (AVG) scores do not include EN results; (iv) $^\dag$: Results from \citet{shi2023language}. 
}
\end{table*}

\end{document}